\documentclass{article}

\usepackage[numbers]{natbib}
\usepackage[final]{tackling_climate_workshop_style}
\usepackage{tackling_climate_workshop_style}
\usepackage[utf8]{inputenc} 
\usepackage[T1]{fontenc}    
\usepackage{hyperref}       
\usepackage{url}            
\usepackage{booktabs}       
\usepackage{amsfonts}       
\usepackage{nicefrac}       
\usepackage{microtype}      

\usepackage{graphics}
\usepackage{graphicx}
\title{AutoML-based Almond Yield Prediction and Projection in California}

\author{%
  Shiheng Duan \\
    Lawrence Livermore National Laboratory \\
    Livermore, CA 94550 \\
    \texttt{duan5@llnl.gov} \\
    \AND
    Shuaiqi Wu \\
    University of California, Davis \\
    Davis, CA 95616 \\
    \texttt{shqwu@ucdavis.edu} \\
    \AND
    Erwan Monier \\
    University of California, Davis \\
    Davis, CA 95616 \\
    \texttt{emonier@ucdavis.edu} \\
    \AND
    Paul Ullrich\\
    University of California, Davis\\
    Davis, CA 95616\\
    \texttt{paullrich@ucdavis.edu}\\
}

\begin{document}

\maketitle

\begin{abstract}
  Almonds are one of the most lucrative products of California, but are also among the most sensitive to climate change. In order to better understand the relationship between climatic factors and almond yield, an automated machine learning framework is used to build a collection of machine learning models. The prediction skill is assessed using historical records. Future projections are derived using 17 downscaled climate outputs. The ensemble mean projection displays almond yield changes under two different climate scenarios, along with two technology development scenarios, where the role of technology development is highlighted. The mean projections and distributions provide insightful results to stakeholders and can be utilized by policymakers for climate adaptation. 
\end{abstract}

\section{Introduction}

California is the most significant almond-producing region in the world. Eighty percent of the almonds consumed globally and all commercially sold in the United States are produced in California. Almonds were California's most profitable agricultural export product and the second-largest commodity overall in 2020 \cite{caCDFAStatistics}. Previous studies have shown that climate factors can influence almond growth \cite{lobell2006impacts, hong2020impacts}, and when these climate factors change due to global warming, the yield of almonds will inevitably suffer. Thus, a precise projection is necessary for both scientific research and climate adaptations. 

Because of the difficulty in simulating the growth process of perennial crops, very few process-based crop models can be utilized to investigate the relationship between climatic variables and almond yields. Data-driven approaches, including machine learning (ML) and deep learning (DL) models, have gained popularity in Earth system modeling since the emergence of artificial intelligence due to their capacity to fit nonlinear functions. Many studies have applied ML and DL models for various tasks, such as hydrology projection, air quality analysis and severe weather forecasting \cite{duan2020using, lv2022meteorology, hill2022new}. The ML and DL models can be designed in an end-to-end manner to examine the relationship between input and output variables. While this strategy is often effective, a proper model architecture is required, which often involves 
specialized knowledge. Automated machine learning (AutoML) frameworks are designed to save these efforts and to facilitate the use of ML models by researchers without a background in computer science, as is frequently the case for climate studies. 

The goal of this study is to build an AutoML framework that creates pipelines from California's climatic variables to estimate almond yields. Section \ref{sec:data} introduces the data sources and AutoML model. In section \ref{sec:prediction}, the model is trained and tested with observational records, and future projections are in section \ref{sec:projection}.

\section{Data and Machine Learning Model} \label{sec:data}
\subsection{GridMET and Almond Yields}
\begin{figure}[h]
    \centering
    \includegraphics[scale=0.23]{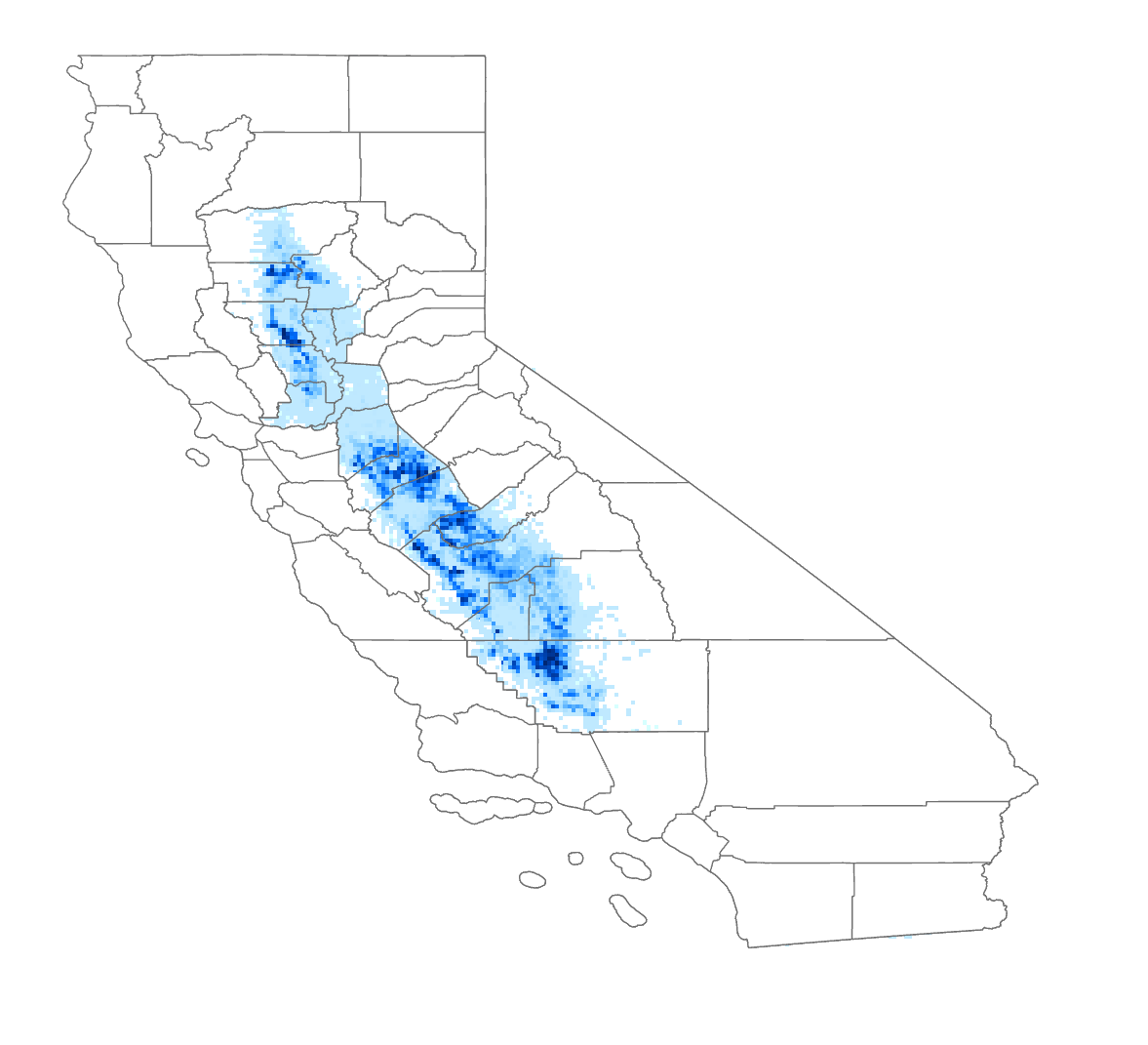}
    \caption{Map of Almond orchards. Darker color represents denser plant area. }
    \label{fig:location}
\end{figure}

Climate variables were derived from the GridMET climate dataset, which provides high-resolution (1/24th degree) daily meteorological data for the contiguous United States from 1979 to the present \cite{abatzoglou2013development}. Instead of simply using monthly or seasonal averaged climate variables, we extracted climate data from GridMET based on the phenology of almond, such as the specific humidity of the bloom period and chill hours of dormancy. Gridded climate variables were averaged or summed based on the historical location of almond orchards from the CropScape geospatial Cropland Data Layer (CDL) product developed by the United States Department of Agriculture, which maintains the consistency between climate variables and almond yield and improves the accuracy of our analysis \cite{USDA-cropland}. Figure \ref{fig:location} depicts the location of almond orchards in California over the past 15 years. Climate data inputs were normalized by removing their mean values and dividing them by their standard deviation. Previous studies have revealed quadratic relationships between climate variables and perennial crops yields in California \cite{lobell2006impacts}. Therefore, we include a total of 13 phenology-climate variables and their squares as features to the ML model. The county-level almond yields (ton/acre) and planted areas (acre) were reported by the California County Agricultural Commissioners and available on the website of USDA (United States Department of Agriculture) for the period of 1980 to 2020 \cite{USDA-NASS-Cal}. To account for advancements in farming practices and technological improvements, we include a linear trend variable for each county. In addition, the county names (16 counties) were used as a categorical predictor in our analysis to represent the differences in static factors such as soil properties and agricultural policies among counties. 

\subsection{MACA}
The Multivariate Adaptive Constructed Analogs (MACA) dataset is used for projection purposes. It includes downscaled products from 20 global climate models from the Coupled Model Intercomparison Project Phase 5 (CMIP5). We subselected the 17 climate models without missing values in our study area. These models have been developed at different modeling centers across the globe, and can account for structural uncertainties such as model design and parameterization choice. The selected models can be found in the supplement. Following the CMIP5 experimental design, the historical period is from 1850 to 2006 \cite{taylor2012overview}. For the period from 2006 to 2020, which overlaps with our observational period and gridMET forcings, both the RCP4.5 and RCP8.5 scenarios are used and compared against observational records.

\subsection{AutoGluon}
Automated machine learning (AutoML) is a type of framework that automates the construction of machine learning pipelines from input data to targets \cite{he2021automl}. It integrates several necessary processes in machine learning applications, including but not limited to: data preprocessing, data augmentation, feature engineering, and hyperparameter tuning. AutoGluon is an AutoML framework that automates ML pipelines on tabular, text and image datasets \cite{erickson2020autogluon}. It has been used for various problems, such as landslide hazards \cite{qi2021autogluon} and drought forecasts \cite{duan2022automl}. In contrast to other AutoML frameworks, AutoGluon places less emphasis on hyperparameter optimization. Instead, it makes use of multi-layer stack ensembling and repeated k-fold bagging, which has been demonstrated to outperform other prominent AutoML frameworks in benchmark tasks \cite{erickson2020autogluon}. 

\section{Prediction Performance} \label{sec:prediction}

The coefficient of determination ($R^2$ score) and root mean squared error (RMSE) are used to quantify the model predictability, and AutoGluon is compared against random forest and linear regression. A 5-fold cross-validation approach is used together with a train-test split method (70\%-30\%) to benchmark the ML models. There are various presets in AutoGluon that specify the hyperparameters. In general, high accuracy corresponds to longer training time and larger models for disk consumption. The choice of presets may affect the model performance, but to maintain simplicity and automation, we only select one preset (`high-quality') in this study. 

As listed in Table \ref{tab:r2prediction}, both the AutoGluon and random forest have relatively lower skill with the train-test split, with the exception of linear regression. This is probably due to the decrease in the number of training samples, since 80\% samples are used as the training set in cross-validation, whereas only 70\% are used for the train-test split. Notably, the AutoGluon model achieves the best performance for both the cross-validation and train-test settings. 

\begin{table}[h]
\centering
\caption{Machine learning model performance. Values in bold represent the best performance. }
\label{tab:r2prediction}
\resizebox{\textwidth}{!}{%
\begin{tabular}{ccc|cc}
\hline
Model                    & Cross-Validation $R^2$ & Cross-Validation RMSE & Train-Test $R^2$ & Train-Test RMSE \\ \hline
AutoGluon & \textbf{0.754}         & \textbf{0.132}        & \textbf{0.740}   & \textbf{0.135}  \\

Linear Regression        & 0.715                  & 0.143                 & 0.724            & 0.139           \\
Random Forest            & 0.611                  & 0.167                 & 0.599            & 0.168           \\ \hline
\end{tabular}%
}
\end{table}

\section{Projection and Climate Adaptation} \label{sec:projection}

\begin{figure}[htp]
    \centering
    \includegraphics[width=\textwidth]{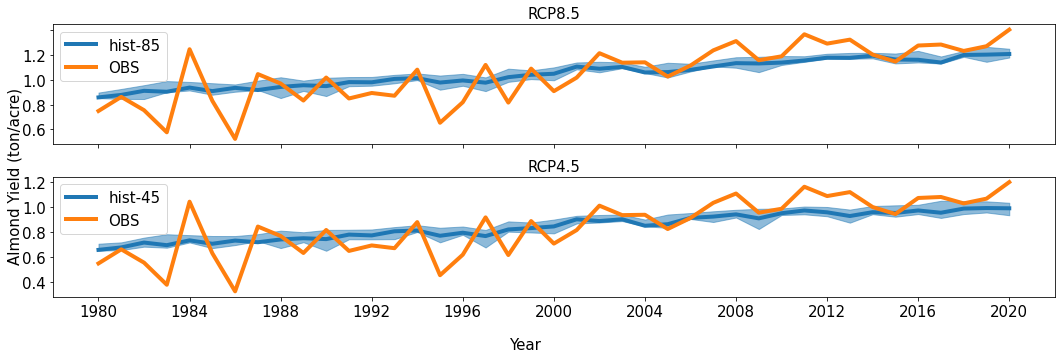}
    \caption{Almond yield observations and historical simulations. The blue line depicts the mean simulations from the selected 17 MACA models, and the blue shading represents the inter-quantile range. Observation is abbreviated to `OBS'. `RCP' projection data is used over the period from 2005 to 2020. }
    \label{fig:hist}
\end{figure}

Projections under various climate change scenarios are produced using the trained AutoGluon model and MACA forcing data. Figure \ref{fig:hist} shows the comparison of the historical simulation with observations. The results from 17 MACA models are generally consistent with one another, with a small inter-quantile range, and are effective at capturing the trend in almond yield. Comparing the ensemble mean simulations, RCP4.5 is slightly closer to observations ($R^2=0.5205$) than RCP8.5 ($R^2=0.5199$). Despite the fact that these values are lower than those in Table \ref{tab:r2prediction}, it is to be expected given that these climate models are only forced by aerosol and greenhouse-gas concentrations and so are only climatologically consistent with observed history. Shorter-term phenomena that can affect yield  (i.e., heat waves or wildfires) are averaged out. 

\begin{figure}[h]
    \centering
    \includegraphics[width=\textwidth]{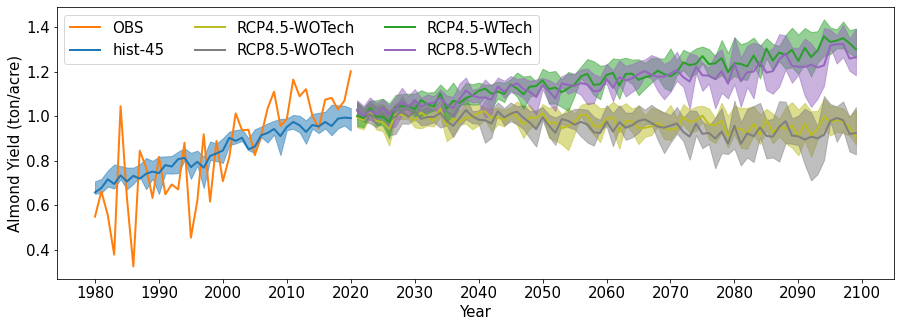}
    \caption{AutoGluon projections under RCP4.5 and RCP8.5 scenarios. Solid lines represent the ensemble mean from the selected MACA simulations, whereas shadings show the inter-quantile range. Observation is abbreviated to `OBS'. }
    \label{fig:proj}
\end{figure}

AutoGluon projections under RCP4.5 and RCP8.5 scenarios are shown in Figure \ref{fig:proj}. In addition to climate change scenarios, we also include two technology scenarios: `WOTech' denotes agricultural technology remaining at its current level as of 2020, whereas `WTech' denotes agricultural technology continuing improving through 2100. Compared with historical simulations, the future projections display higher uncertainty, with larger inter-quantile ranges, likely due to the discrepancy among climate models. It is evident by comparing the various technology scenarios within the same climatic scenario that technological advancement is necessary to maintain historical trends in annual almond yield growth. This finding highlights the important role of technological advancement in climate adaptation. On the other hand, almond yield will decline in both the RCP4.5 and RCP8.5 scenarios after peak technological advancement. As for the climate change scenarios, the RCP8.5 shows a slightly greater decline in yields compared with RCP4.5, likely due to high temperatures under this scenario.

\section{Conclusions and Future Work}

The use of deep learning and machine learning techniques to answer climate change-related problems is growing in popularity. Not all tasks, however, can offer sufficient data samples for complex neural networks. In this study, we demonstrate how an AutoML framework can predict and project California's almond yield. As an ensemble ML model, AutoGluon achieves better performance compared with single ML models, without requiring more data samples or computational resources. It also enables us to analyze the relative importance of input variables using, i.e., a permutation-based method, from which the total precipitation during the bloom period (February-Mid March) is shown as the most important variable. These results are not shown here, as they are unrelated to our projection analysis. 

Our projection results highlight the critical role that technology plays in climate adaptation. Increases in almond yield require further technological developments; in the absence of further development, climatological factors will lead to a decrease in yields. However, the cost of developing new techniques is not taken into consideration in our analysis. From the perspective of stakeholders, it could be promising to incorporate our projection results within a socioeconomic model to gain insight into the balance between investment cost and expected output. 

\section{Acknowledgment}

We would like to thank computational resources funded by the Walter and Margaret Milton Graduate Atmospheric Science Award at University of California, Davis. The workshop presentation is funded by the Lawrence Livermore National Laboratory Postdoc Development Program. Part of this work was performed under the auspices of the U.S. Department of Energy by Lawrence Livermore National Laboratory under Contract DE-AC52-07NA27344.

{\small
\bibliographystyle{unsrtnat}
\bibliography{ref}

\begin{thebibliography}{14}
\providecommand{\natexlab}[1]{#1}
\providecommand{\url}[1]{\texttt{#1}}
\expandafter\ifx\csname urlstyle\endcsname\relax
  \providecommand{\doi}[1]{doi: #1}\else
  \providecommand{\doi}{doi: \begingroup \urlstyle{rm}\Url}\fi

\bibitem[caC()]{caCDFAStatistics}
{C}{D}{F}{A} - {S}tatistics --- cdfa.ca.gov.
\newblock \url{https://www.cdfa.ca.gov/Statistics}.
\newblock [Accessed 14-Aug-2022].

\bibitem[Lobell et~al.(2006)Lobell, Field, Cahill, and
  Bonfils]{lobell2006impacts}
David~B Lobell, Christopher~B Field, Kimberly~Nicholas Cahill, and Celine
  Bonfils.
\newblock Impacts of future climate change on california perennial crop yields:
  Model projections with climate and crop uncertainties.
\newblock \emph{Agricultural and Forest Meteorology}, 141\penalty0
  (2-4):\penalty0 208--218, 2006.

\bibitem[Hong et~al.(2020)Hong, Mueller, Burney, Zhang, AghaKouchak, Moore,
  Qin, Tong, and Davis]{hong2020impacts}
Chaopeng Hong, Nathaniel~D Mueller, Jennifer~A Burney, Yang Zhang, Amir
  AghaKouchak, Frances~C Moore, Yue Qin, Dan Tong, and Steven~J Davis.
\newblock Impacts of ozone and climate change on yields of perennial crops in
  california.
\newblock \emph{Nature Food}, 1\penalty0 (3):\penalty0 166--172, 2020.

\bibitem[Duan et~al.(2020)Duan, Ullrich, and Shu]{duan2020using}
Shiheng Duan, Paul Ullrich, and Lele Shu.
\newblock Using convolutional neural networks for streamflow projection in
  california.
\newblock \emph{Frontiers in Water}, page~28, 2020.

\bibitem[Lv et~al.(2022)Lv, Tian, Luo, Liu, Bai, Zhao, Lin, Zhao, Guo, Xiao,
  et~al.]{lv2022meteorology}
Yunqian Lv, Hezhong Tian, Lining Luo, Shuhan Liu, Xiaoxuan Bai, Hongyan Zhao,
  Shumin Lin, Shuang Zhao, Zhihui Guo, Yifei Xiao, et~al.
\newblock Meteorology-normalized variations of air quality during the covid-19
  lockdown in three chinese megacities.
\newblock \emph{Atmospheric Pollution Research}, page 101452, 2022.

\bibitem[Hill et~al.(2022)Hill, Schumacher, and Jirak]{hill2022new}
Aaron~J Hill, Russ~S Schumacher, and Israel Jirak.
\newblock A new paradigm for medium-range severe weather forecasts:
  probabilistic random forest-based predictions.
\newblock \emph{arXiv preprint arXiv:2208.02383}, 2022.

\bibitem[Abatzoglou(2013)]{abatzoglou2013development}
John~T Abatzoglou.
\newblock Development of gridded surface meteorological data for ecological
  applications and modelling.
\newblock \emph{International Journal of Climatology}, 33\penalty0
  (1):\penalty0 121--131, 2013.

\bibitem[USD({\natexlab{a}})]{USDA-cropland}
{C}{D}{F}{A} - {S}tatistics --- cdfa.ca.gov.
\newblock
  \url{https://www.nass.usda.gov/Research_and_Science/Cropland/sarsfaqs2.php},
  {\natexlab{a}}.
\newblock [Accessed 14-Aug-2022].

\bibitem[USD({\natexlab{b}})]{USDA-NASS-Cal}
{C}{D}{F}{A} - {S}tatistics --- cdfa.ca.gov.
\newblock
  \url{https://www.nass.usda.gov/Statistics_by_State/California/Publications/AgComm/},
  {\natexlab{b}}.
\newblock [Accessed 14-Aug-2022].

\bibitem[Taylor et~al.(2012)Taylor, Stouffer, and Meehl]{taylor2012overview}
Karl~E Taylor, Ronald~J Stouffer, and Gerald~A Meehl.
\newblock An overview of cmip5 and the experiment design.
\newblock \emph{Bulletin of the American meteorological Society}, 93\penalty0
  (4):\penalty0 485--498, 2012.

\bibitem[He et~al.(2021)He, Zhao, and Chu]{he2021automl}
Xin He, Kaiyong Zhao, and Xiaowen Chu.
\newblock Automl: A survey of the state-of-the-art.
\newblock \emph{Knowledge-Based Systems}, 212:\penalty0 106622, 2021.

\bibitem[Erickson et~al.(2020)Erickson, Mueller, Shirkov, Zhang, Larroy, Li,
  and Smola]{erickson2020autogluon}
Nick Erickson, Jonas Mueller, Alexander Shirkov, Hang Zhang, Pedro Larroy,
  Mu~Li, and Alexander Smola.
\newblock Autogluon-tabular: Robust and accurate automl for structured data.
\newblock \emph{arXiv preprint arXiv:2003.06505}, 2020.

\bibitem[Qi et~al.(2021)Qi, Xu, and Xu]{qi2021autogluon}
Wenwen Qi, Chong Xu, and Xiwei Xu.
\newblock Autogluon: A revolutionary framework for landslide hazard analysis.
\newblock \emph{Natural Hazards Research}, 1\penalty0 (3):\penalty0 103--108,
  2021.

\bibitem[Duan(2022)]{duan2022automl}
Shiheng Duan.
\newblock Automl-based drought forecast with meteorological variables.
\newblock \emph{arXiv preprint arXiv:2207.07012}, 2022.

\end{thebibliography}
}
\end{document}